%% file: main.tex
\documentclass[10pt,twocolumn,letterpaper]{article}
\usepackage[pagenumbers]{cvpr}
\input{preamble}
\definecolor{cvprblue}{rgb}{0.21,0.49,0.74}
\usepackage[breaklinks,colorlinks,allcolors=cvprblue]{hyperref}
\def\confName{CVPR}
\def\confYear{2027}
\title{FluentAct: Efficient Action Continuations for Asynchronous Vision-Language-Action Control}
\newcommand{\fluentactauthors}{%
\parbox{\dimexpr\textwidth-2\tabcolsep\relax}{\centering
  {\fontsize{11}{15}\selectfont
  Shuo Bao\textsuperscript{1,*}\quad
  Ming Shang\textsuperscript{2,*}\quad
  Yuchen Huang\textsuperscript{3,*}\quad
  Hongjin Niu\textsuperscript{4,*}\\
  Han Yu\textsuperscript{5}\quad
  Yilong Qiu\textsuperscript{5}\quad
  Yifan Wang\textsuperscript{6}\quad
  Shuo Tian\textsuperscript{7}\\
  Jiangtao Wang\textsuperscript{8}\quad
  Hongyu Wang\textsuperscript{9}\quad
  Xinzhou Wang\textsuperscript{10,\ensuremath{\dagger}}\quad
  Fuchun Sun\textsuperscript{11,\ensuremath{\dagger}}\par}
  \vspace{5pt}
  {\fontsize{8.5}{11}\selectfont
  \textsuperscript{1}Peking University\quad
  \textsuperscript{2}Beihang University\quad
  \textsuperscript{3}University of Science and Technology Beijing\\
  \textsuperscript{4}Xi'an Jiaotong University\quad
  \textsuperscript{5}Beijing Institute of Technology\quad
  \textsuperscript{6}Harbin Institute of Technology\\
  \textsuperscript{7}Beijing University of Chemical Technology\quad
  \textsuperscript{8}Monash University\quad
  \textsuperscript{9}Robot Forge\\
  \textsuperscript{10}Tongji University\quad
  \textsuperscript{11}Tsinghua University\par}
  \vspace{4pt}
  {\fontsize{9}{11}\selectfont
  \textsuperscript{*}Equal contribution.\qquad
  \textsuperscript{\ensuremath{\dagger}}Corresponding authors.\par}
}}
\author{\fluentactauthors}
\hypersetup{
  pdftitle={FluentAct: Efficient Action Continuations for Asynchronous Vision-Language-Action Control},
  pdfauthor={Shuo Bao, Ming Shang, Yuchen Huang, Hongjin Niu, Han Yu, Yilong Qiu, Yifan Wang, Shuo Tian, Jiangtao Wang, Hongyu Wang, Xinzhou Wang, Fuchun Sun}
}
\begin{document}
\maketitle
\raggedbottom
\input{sec/0_abstract}
\input{sec/1_introduction}
\input{sec/2_related_work}
\input{sec/3_method}
\input{sec/4_experiments}
\input{sec/5_conclusion}
\clearpage
\section*{LLM Usage Statement}
In this paper, we employ Large Language Models (LLMs)
solely for minor linguistic refinement during the manuscript
preparation stage, such as correcting grammatical errors.
None of the technical content, implementation details, or
experimental results were generated by LLMs.
{\small
\bibliographystyle{ieeenat_fullname}
\bibliography{references}
}
\end{document}

%% file: preamble.tex
\usepackage[T1]{fontenc}
\usepackage{graphicx}
\usepackage{amsmath,amssymb}
\usepackage{booktabs,array,tabularx,multirow}
\usepackage{enumitem}
\usepackage{siunitx}
\graphicspath{{fig/}}

\newcommand{\sg}{\operatorname{sg}}
\newcommand{\LN}{\operatorname{LN}}
\newcommand{\Attn}{\operatorname{Attn}}
\newcommand{\E}{\mathbb{E}}

%% file: sec/0_abstract.tex
\begin{abstract}
Asynchronous vision-language-action (VLA) policies generate
new action chunks while robots continue executing committed
commands. Because the robot moves during inference, a chunk
generated from an earlier observation can introduce abrupt
command changes at handoff. To address this problem, we introduce FluentAct, a
framework for efficient action continuation that improves
command continuity at handoff.
Instruction-aware history and retrieved memory provide task
context, while a latent rollout of committed actions constructs
the handoff context.
An ordered Haar generator predicts pairwise averages and
conditional differences with two forward evaluations of a shared
action expert.
During offline training, a frozen reference samples candidates
under the same request and committed prefix.
We rank these candidates by command continuity and train the
policy with a reference-relative flow-energy objective.
The trained policy generates continuations directly without
online candidate ranking.
FluentAct improves task completion over BehaviorVLA on CALVIN,
RoboTwin~2.0 Hard, and six real-world manipulation tasks,
achieving a CALVIN average sequence length of 4.52.
On LIBERO, it achieves 99.1\% success and reduces handoff
command jump and trend mismatch by 21.3\% and 20.7\%,
respectively, relative to the variant without preference training.
In a separate timing study in unpaced LIBERO simulation on
an RTX~4090, its median request-to-handoff latency is
153\,ms, 32.5\% lower than native $\pi_{0.5}$.
\end{abstract}

%% file: sec/1_introduction.tex
\section{Introduction}
\label{sec:introduction}


\begin{figure}[t]
\centering
\includegraphics[width=\linewidth]{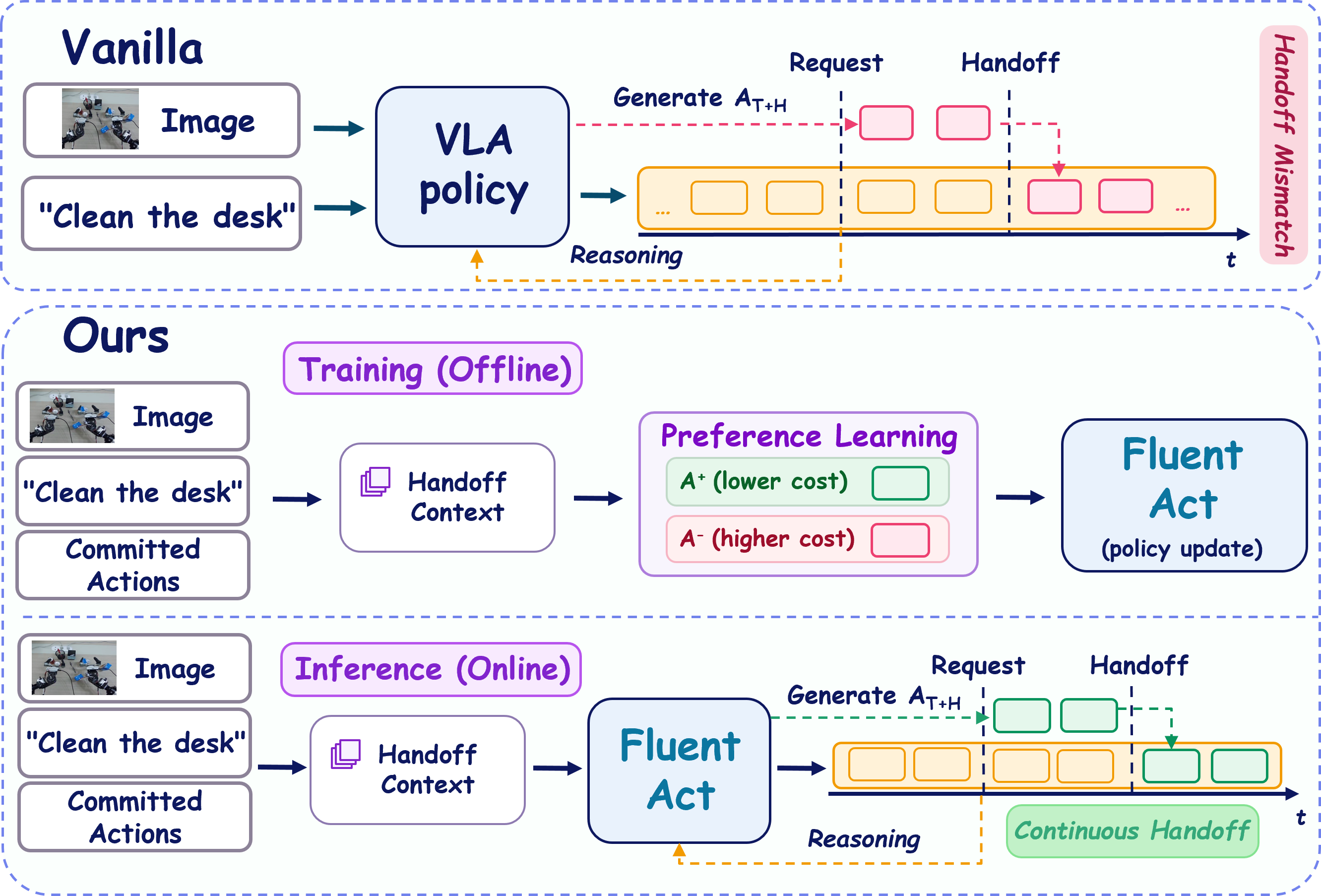}
\caption{Asynchronous action handoffs in vanilla VLA and FluentAct. Top: a new chunk can introduce an abrupt command change at handoff.
Middle: offline preference learning compares continuations under
the same request and committed prefix.
Bottom: FluentAct directly generates a continuation conditioned
on the handoff context.}
\label{fig:motivation}
\end{figure}

Action chunking enables vision-language-action (VLA) policies to predict multiple control commands in a single inference step~\cite{zhao2023act,black2024pi0,black2025pi05}. To support continuous robot control, asynchronous execution overlaps action generation with execution: while the policy generates a new action chunk, the robot continues executing commands from the previous chunk~\cite{black2025real}. However, this introduces a critical handoff problem. By the time a new chunk takes control, the robot has already executed additional commands beyond the observation that triggered the request. The incoming chunk must therefore not only be generated before the handoff deadline, but also connect consistently to the motion already committed by the robot.

Failure to satisfy these requirements can interrupt continuous control or introduce abrupt command changes at chunk boundaries, potentially disrupting contact-sensitive manipulation and long-horizon task execution. For example, when a robot arm carries an object toward a placement region, two candidate continuations may both approach the target, yet one preserves the outgoing motion trend while the other introduces a sudden change at handoff (Fig.~\ref{fig:motivation}). Existing approaches address asynchronous execution by conditioning generation on committed actions or reducing the computational cost of action generation~\cite{black2025real,black2025training,liu2026learning,wang2026real,wang2025onedp,su2025freqpolicy}. While these strategies improve execution awareness or generation efficiency, they do not explicitly optimize the relative handoff quality of alternative continuations under the same request and committed prefix. In particular, standard imitation supervision does not distinguish between plausible continuations according to how well they connect to the outgoing commands.



To address this problem, we introduce \textbf{FluentAct}, a framework for efficient and command-consistent action continuation in asynchronous VLA control. Building on the causal history representation and retrieved memory of BehaviorVLA~\cite{hu2026from}, FluentAct incorporates instruction-aware behavior context and constructs a handoff representation by rolling committed actions forward through a frozen latent history transition. Conditioned on this representation, a shared action expert generates the next continuation through an ordered Haar parameterization, first predicting pairwise action averages and then their conditional differences. This design reconstructs the action continuation with only two sequential evaluations of the shared action expert, enabling efficient generation during ongoing execution.


Beyond handoff conditioning, FluentAct explicitly learns which continuations better preserve the outgoing motion. During offline training, a frozen reference policy generates alternative continuations under the same request and committed prefix. We compare their command jump and motion-trend mismatch at the handoff boundary, and construct preference pairs subject to reference-energy compatibility constraints. A reference-relative flow-energy objective then encourages the policy to favor lower-cost continuations, while demonstration replay preserves task supervision. Unlike inference-time candidate selection, this preference is learned directly by the action generator, allowing FluentAct to produce continuations without additional online ranking.

Experiments across CALVIN, RoboTwin~2.0 Hard, LIBERO, and six real-world manipulation tasks demonstrate improvements in task completion and handoff continuity. On LIBERO, continuation preference learning reduces command jump and trend mismatch by 21.3\% and 20.7\%, respectively, while a separate timing study reports a median request-to-handoff latency of 153\,ms.
Our main contributions are as follows:
\begin{itemize}
    \item We introduce FluentAct, a framework for efficient
    asynchronous VLA control. Its handoff-conditioned generator
    combines instruction-aware behavior context with a latent
    rollout of committed commands. Ordered Haar generation
    produces continuations in two forward evaluations of a
    shared action expert.

    \item We introduce offline continuation preference learning
    from reference-policy candidates conditioned on the same
    request and committed prefix. A reference-relative flow-energy
    objective trains the policy to favor command continuity
    and also applies to the native flow generators of
    $\pi_{0.5}$-FM, Legato, and RTC.

    \item We evaluate the complete framework on three simulation
    benchmarks and six real-world manipulation tasks.
    FluentAct achieves 69.3\% mean success on the real-world
    tasks, compared with 60.2\% for BehaviorVLA.
    In a separate timing study in unpaced LIBERO simulation
    on an RTX~4090, its median request-to-handoff latency is
    153\,ms, 32.5\% lower than native $\pi_{0.5}$.
\end{itemize}

%% file: sec/2_related_work.tex
\section{Related Work}
\label{sec:related}

\paragraph{Action generation.}
OpenVLA predicts discrete action tokens~\cite{kim2024openvla},
while $\pi_0$ generates continuous chunks through flow
matching~\cite{black2024pi0}. BehaviorVLA adds causal
visuomotor history and retrieved behavior
memory~\cite{hu2026from}.
Consistency Policy and OneDP accelerate diffusion sampling
through distillation~\cite{prasad2024consistencypolicy,wang2025onedp}.
For flow policies, conditional optimal-transport couplings
support few-step sampling~\cite{sochopoulos2025fast}, while
FreqPolicy uses frequency consistency for one-step
generation~\cite{su2025freqpolicy}.

\paragraph{Asynchronous continuation.}
RTC conditions on committed actions through inference-time
inpainting~\cite{black2025real}. Training-time action
conditioning, Legato, and REMAC learn prefix-conditioned
generation~\cite{black2025training,liu2026learning,wang2026real}.
VLASH extrapolates explicit robot states from issued
actions~\cite{tang2025vlash}, whereas FluentAct rolls a learned
history latent through the committed prefix.

\paragraph{Preference learning for flow models.}
DPO learns preferences relative to a reference
policy~\cite{rafailov2023direct}. FlowPRO obtains VLA preferences
through intervention and rollback~\cite{wu2026flowpro}.
DEFLECT constructs fresh/stale counterfactual preferences
from a frozen reference and uses reference-relative
flow-matching scores~\cite{zhu2026deflect}.
FluentAct instead ranks candidates generated under the same
request and committed prefix by handoff command costs.

%% file: sec/3_method.tex
\section{Method}
\label{sec:method}

\subsection{Problem Formulation}
\label{sec:handoff-setup}
In asynchronous VLA control, action generation overlaps with
execution~\cite{black2025training}.
At request step $r$, the policy receives observation $o_r$,
instruction $\ell$, and causal history $\mathcal H_r$.
During inference, the robot executes a committed prefix
$U_{r,k}=(u_r,\ldots,u_{r+k-1})$ before the continuation
$A=(a_0,\ldots,a_{H-1})$ takes over at step $r+k$.
Because the robot moves beyond the request-time observation,
the new continuation may introduce abrupt command changes
at takeover.
Prefix conditioning supplies execution context, but imitation
alone does not explicitly compare alternative continuations
at the same handoff.

FluentAct combines handoff-conditioned action generation
with offline continuation preference learning.
Instruction-aware history and retrieved memory provide
task-relevant behavior context
(\ref{sec:behavior-context}).
A handoff-conditioned Haar generator incorporates the
committed prefix to predict the continuation in two sequential
evaluations of a shared action expert
(\ref{sec:handoff-context}).
During offline training, we compare candidates under the
same request and committed prefix, using command jump
and command-trend mismatch at handoff to construct
continuation preferences
(\ref{sec:candidate-filtering}).
A reference-relative flow-energy objective transfers these
preferences to the action generator, which directly produces
continuations at deployment without online candidate
selection (\ref{sec:flow-energy-preference}).

\begin{figure*}[t]
\centering
\includegraphics[width=\textwidth]{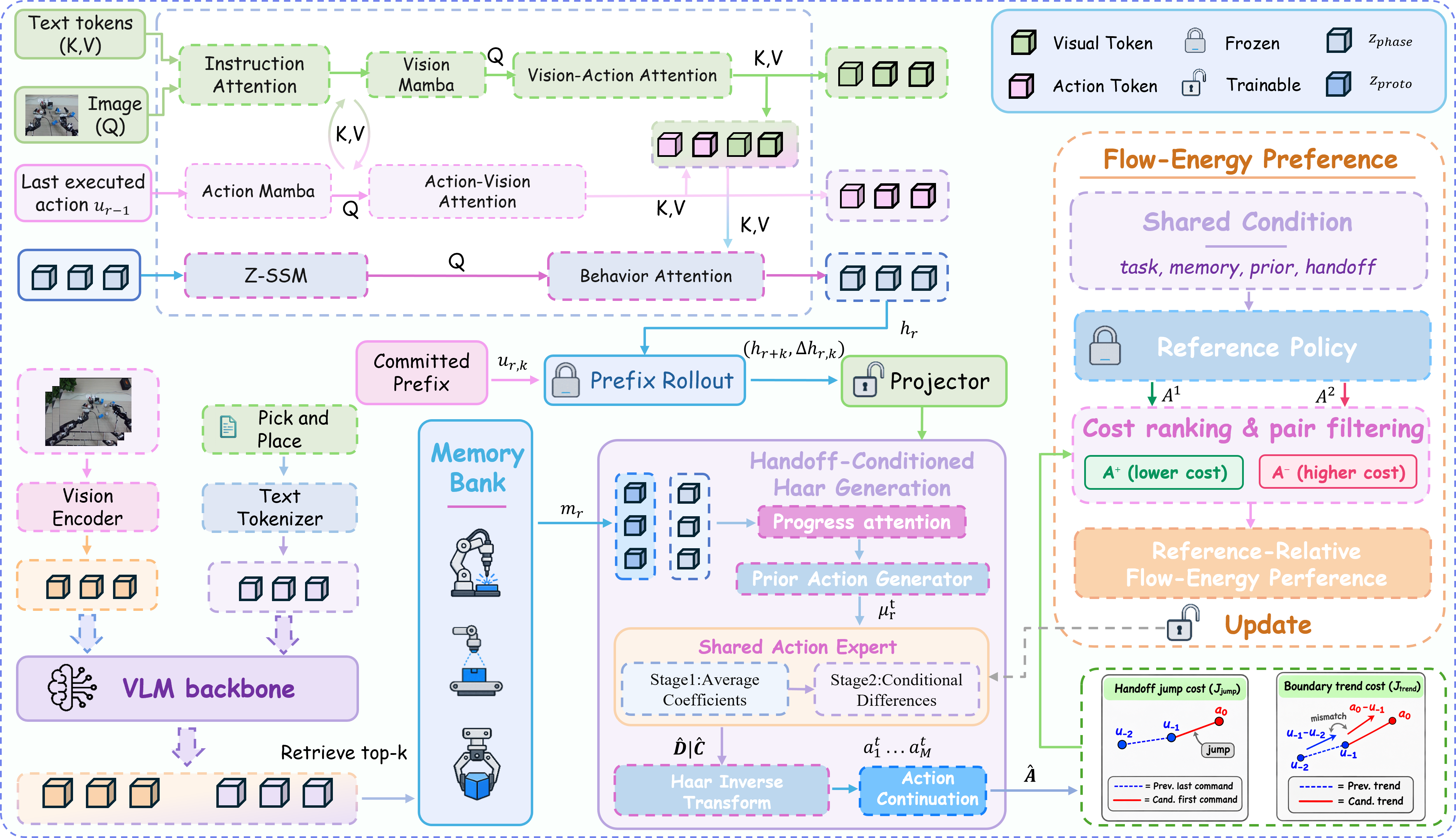}
\caption{Overview of FluentAct.
Instruction-aware behavior context and a latent rollout of
committed commands condition ordered Haar generation.
A shared action expert predicts average coefficients followed
by conditional differences.
During offline training, handoff costs rank reference-policy
continuations for preference learning.}
\label{fig:framework}
\end{figure*}

\subsection{Instruction-Aware Behavior and Memory}
\label{sec:behavior-context}

We build on the causal visuomotor history and memory structure
of BehaviorVLA~\cite{hu2026from}, placing instruction conditioning
before spatial pooling and temporal aggregation.
For visual tokens $V_t\in\mathbb R^{N_v\times d}$ and language
features $L\in\mathbb R^{N_\ell\times d_\ell}$, let
$L_b=\LN(LW_\ell)$ with
$W_\ell\in\mathbb R^{d_\ell\times d}$.
Visual queries attend to instruction keys and values:
\begin{equation}
\widetilde V_t
=V_t+\Attn\big(\LN(V_t),L_b,L_b;M_\ell\big),
\label{eq:visual_query}
\end{equation}
where $M_\ell$ masks instruction padding.
Vision and action Mamba branches~\cite{gu2023mamba},
reciprocal visuomotor attention, and a recurrent behavior
stream form the behavior encoder $\mathcal B_\phi$.
It aggregates instruction-conditioned visual observations
up to step $r$ and executed commands before step $r$
into the behavior state $z_r$.

The memory bank stores trajectory-level retrieval keys and
pooled behavior representations from demonstrations.
A VLM query retrieves the top-$K$ entries, and weighted pooling
of their values produces the memory state $m_r$.
An action-prior decoder $\mathcal D_\omega$ combines the current
behavior state $z_r$ and retrieved memory $m_r$ to predict
an action prior $\mu_r\in\mathbb R^{H\times d_a}$ for the
current task phase, where $d_a$ is the action dimension.
The prior and behavior context condition both stages of
Haar generation.
The supplementary material specifies the encoder, retrieval
interface, and pretraining objectives.

\subsection{Handoff-Conditioned Haar Generation}
\label{sec:handoff-context}

\paragraph{Latent handoff context.}
The behavior state $z_r$ summarizes execution history at the
request time, while the continuation starts after the committed
prefix has executed. We therefore use the known commands to
predict a latent context for the upcoming handoff.
Starting from $h_r=z_r$, we apply a history transition
$R_{\mathrm{act}}$, pretrained to predict the next history
latent from the current latent and executed command.
The transition remains frozen. We retain both its predicted
handoff state and the change from the request state, then
project them into the action expert's conditioning space:
\begin{align}
h_{j+1}
&=R_{\mathrm{act}}(h_j,u_j),
\quad j=r,\ldots,r+k-1,
\label{eq:roll_step}\\
\Delta h_{r,k}
&=h_{r+k}-h_r,
\label{eq:roll_delta}\\
g_\psi
&=P_\psi\big(
[\LN(\sg(h_{r+k}));
\LN(\sg(\Delta h_{r,k}))]\big).
\label{eq:handoff-context}
\end{align}
Here $\sg$ stops gradients, brackets denote concatenation,
and $P_\psi$ is trainable.

The generator combines this handoff context with the
request-time task context from Sec.~\ref{sec:behavior-context}.
We collect its inputs as
$\chi=(\Phi_r,z_r,m_r,\mu_r,h_{r+k},\Delta h_{r,k})$,
where $\Phi_r$ denotes the visual-language backbone features.
Each velocity predictor below applies $P_\psi$ internally,
and its parameters $\theta$ include $\psi$.
During preference learning, the current policy and frozen
reference receive the same unprojected inputs $\chi$ and
apply their own projectors, with the reference projector frozen.

\paragraph{Pairwise Haar coordinates.}
To organize continuation generation into two stages, we
represent each adjacent action pair by its average component
and within-pair difference. A one-level Haar
transform~\cite{mallat1989wavelet} provides this decomposition.
For an even prediction horizon $H$, the coefficients are
\begin{equation}
C_i=\frac{a_{2i}+a_{2i+1}}{\sqrt2},
\qquad
D_i=\frac{a_{2i}-a_{2i+1}}{\sqrt2},
\label{eq:haar_pair}
\end{equation}
for $i=0,\ldots,H/2-1$.
The scaled averages $C$ form a pairwise motion scaffold,
while $D$ retains the variation within each pair.
Let $W_H$ denote the corresponding orthogonal transform.
Together, the two coefficient groups reconstruct the chunk
exactly:
\begin{equation}
\begin{bmatrix}
C\\
D
\end{bmatrix}
=W_HA,
\qquad
A=W_H^\top
\begin{bmatrix}
C\\
D
\end{bmatrix}
\end{equation}
The action prior uses the same coordinates,
$[\mu_r^C;\mu_r^D]=W_H\mu_r$.

\paragraph{Ordered generation with a shared expert.}
The generator first predicts the scaffold $C$ and then
predicts the differences $D$ conditioned on that scaffold.
A shared action expert implements both stages, with stage
embeddings and coefficient masks distinguishing the
predictors $v_\theta^C$ and $v_\theta^D$.
Each stage uses a conditional flow predictor.
For target coefficients $y$ and Gaussian noise $\epsilon$,
we interpolate between data at $\tau=0$ and noise at $\tau=1$:
\begin{equation}
x_\tau=(1-\tau)y+\tau\epsilon,
\qquad
v^\star=\epsilon-y.
\label{eq:flow_path}
\end{equation}
At inference, each stage takes one explicit Euler step from
$\tau=1$ to $\tau=0$.
Starting from noises $\epsilon_C$ and $\epsilon_D$, the
two stages generate a continuation as
\begin{align}
\widehat C
&=\epsilon_C-v_\theta^C(\epsilon_C,1;\chi),
\label{eq:coarse_stage}\\
\widehat D
&=\epsilon_D-v_\theta^D
(\epsilon_D,1;\widehat C,\chi),
\label{eq:detail_stage}\\
\widehat A
&=W_H^
\top\begin{bmatrix}
    \widehat C \\ \widehat D
\end{bmatrix}
\nonumber
\end{align}
This requires two sequential evaluations of the shared
action expert, followed by the fixed inverse Haar transform.

To train this generator from demonstrations, we transform
each target continuation $A^\star$ into
$[C^\star;D^\star]=W_HA^\star$.
The interpolants $x_\tau^C$ and $x_\tau^D$ follow
Eq.~\eqref{eq:flow_path} with these target coefficients.
The difference predictor is conditioned on the scaffold
$\widehat C$ generated by Eq.~\eqref{eq:coarse_stage}:
\begin{align}
\widehat v_C
&=v_\theta^C(x_\tau^C,\tau;\chi),
\nonumber\\
\widehat v_D
&=v_\theta^D
(x_\tau^D,\tau;\sg(\widehat C),\chi).
\nonumber
\end{align}
Stop-gradient blocks the difference loss from propagating
through this conditioning input.
We minimize
\begin{equation}
\begin{split}
\mathcal L_{\mathrm{struct}}
=\tfrac12\big(&
\|\widehat v_C-(\epsilon_C-C^\star)\|_2^2\\
&+\|\widehat v_D-(\epsilon_D-D^\star)\|_2^2
\big).
\end{split}
\label{eq:struct_loss}
\end{equation}
The supplementary material describes the stage inputs,
coefficient masks, and prior projections.

\subsection{Continuation Preference Construction}
\label{sec:candidate-filtering}

We construct continuation preferences by comparing candidates
at the same handoff. The frozen reference generates two
candidates using different sampling noises while keeping the
request, behavior context, and committed prefix fixed.
For prefixes with $k\ge2$, let $u_{-1}=u_{r+k-1}$ and
$u_{-2}=u_{r+k-2}$ denote the final two outgoing commands.
For candidate ranking, we penalize the magnitude of the
boundary increment, $a_0-u_{-1}$, and its mismatch with
the last outgoing increment, $u_{-1}-u_{-2}$:
\begin{align}
J_{\mathrm{jump}}(A)
&=\|a_0-u_{-1}\|_W^2,
\nonumber\\
J_{\mathrm{trend}}(A)
&=\|(a_0-u_{-1})-(u_{-1}-u_{-2})\|_W^2,
\nonumber\\
J(A)
&=\lambda_0J_{\mathrm{jump}}(A)
+\lambda_1J_{\mathrm{trend}}(A).
\label{eq:continuation-ranking}
\end{align}
Here $\|x\|_W^2=x^\top Wx$, with nonnegative diagonal
channel weights $W$.
Costs are computed in de-normalized command units.
Equivalently, in normalized coordinates, each channel weight
is multiplied by the square of its de-normalization scale.

A candidate is eligible if all values in its real action
channels are finite and have absolute normalized values
at most $a_{\max}$.
For two eligible candidates, let $A^+$ and $A^-$ denote
the lower-cost and higher-cost continuations, respectively.
We retain pairs with a command-cost gap exceeding $\gamma$
and a reference-energy difference bounded by $\delta_{\mathrm{ref}}$:
\begin{equation}
\begin{gathered}
d=J(A^-)-J(A^+)>\gamma,\\
|E_{\mathrm{ref}}(A^-\mid\chi)
-E_{\mathrm{ref}}(A^+\mid\chi)|
\le\delta_{\mathrm{ref}},
\end{gathered}
\label{eq:pair-acceptance}
\end{equation}
where $E_{\mathrm{ref}}$ is the frozen reference's regression
energy defined in \ref{sec:flow-energy-preference}.

For each accepted pair, we weight the pairwise preference
loss by $q=d/(d+\max(\gamma,10^{-6}))$.
Only accepted pairs contribute to the preference and
preferred-candidate regression terms.
Demonstration replay uses all requests, including those
without an accepted pair.

\subsection{Reference-Relative Flow-Energy Learning}
\label{sec:flow-energy-preference}

We train the generator to favor the continuations selected
by the command-cost comparison in
Sec.~\ref{sec:candidate-filtering}.
Each candidate is scored by the velocity-regression energy
of its Haar coefficients. Lower energy indicates a closer
fit to that candidate's regression targets.
For $[C;D]=W_HA$, we score the averages $C$ and the
differences $D$ conditioned on the same candidate's $C$:
\begin{align}
E_\theta^C(C\mid\chi)
&=\E_{\tau,\epsilon_C}
\|v_\theta^C(x_\tau^C,\tau;\chi)
-(\epsilon_C-C)\|_2^2,
\nonumber\\
E_\theta^D(D\mid C,\chi)
&=\E_{\tau,\epsilon_D}
\big\|v_\theta^D(x_\tau^D,\tau;C,\chi)
\nonumber\\
&\hspace{22mm}-(\epsilon_D-D)\big\|_2^2,
\label{eq:stage_energy}\\
E_\theta(A\mid\chi)
&=\tfrac12\big[
E_\theta^C(C\mid\chi)
+E_\theta^D(D\mid C,\chi)
\big].
\label{eq:energy}
\end{align}
The interpolants follow Eq.~\eqref{eq:flow_path} with the
candidate coefficients as targets. Squared errors are summed
over coefficient positions and real action channels.
These energies provide surrogate compatibility
scores~\cite{wu2026flowpro}.
Evaluating the same energy with the frozen reference gives
$E_{\mathrm{ref}}$.

For each request, we estimate the energies using one sampled
flow time and stage-specific Gaussian noises, shared across
both candidates and both policies.
These scoring draws are separate from the noises used for
candidate generation. The reference-energy estimates are
reused for pair filtering and preference optimization.
To transfer the command-cost preference to the generator,
we compare the current policy's energy gap with the frozen
reference's gap, following the reference-relative comparison
in DPO~\cite{rafailov2023direct}:
\begin{equation}
\begin{split}
\Delta_\theta
={}&\big[E_\theta(A^-\mid\chi)
-E_\theta(A^+\mid\chi)\big]\\
&-\big[E_{\mathrm{ref}}(A^-\mid\chi)
-E_{\mathrm{ref}}(A^+\mid\chi)\big].
\end{split}
\label{eq:dpo}
\end{equation}
A positive $\Delta_\theta$ means that the energy gap has
shifted toward $A^+$ relative to the reference.

The training objective combines this pairwise comparison
with regression on preferred candidates and demonstration
replay:
\begin{equation}
\begin{split}
\mathcal L={}&
\E_{\mathrm{accepted}}\big[
-q\log\sigma(\beta\Delta_\theta)\\
&\qquad+\lambda_+E_\theta(A^+\mid\chi)\big]\\
&+\lambda_{\mathrm{rep}}\mathcal L_{\mathrm{replay}}.
\end{split}
\label{eq:continuation-preference-objective}
\end{equation}
Here $\sigma$ is the logistic sigmoid and $\beta>0$
controls the scale of the pairwise objective.
The weight $q$ applies to the pairwise term, while the
preferred-candidate term provides regression supervision
for $A^+$.
Both candidate terms are averaged over accepted pairs in
the training batch. If no pair is accepted, they contribute
zero. Demonstration replay averages
$E_\theta(A^\star\mid\chi)$ over all requests in the batch,
retaining demonstration supervision.

We update the shared action expert and its action, time,
stage, and continuation projections, including $P_\psi$.
The vision-language backbone, behavior encoder, memory bank,
action-prior decoder $\mathcal D_\omega$, history transition
$R_{\mathrm{act}}$, and reference policy remain frozen.

\paragraph{Training and deployment.}
Training proceeds through behavior pretraining and memory
construction, action-policy imitation, handoff-conditioned
Haar imitation, and continuation preference learning.
Candidate sampling, pair filtering, and reference-relative
optimization are performed offline.
At deployment, FluentAct processes the committed prefix
to obtain the handoff context.
It generates $\widehat C$, predicts $\widehat D$ conditioned
on $\widehat C$, and reconstructs the continuation through
the inverse Haar transform.

%% file: sec/4_experiments.tex

\section{Experiments}
\label{sec:experiments}

We evaluate FluentAct on long-horizon and bimanual manipulation
benchmarks and real-world tasks, then examine handoff quality,
execution efficiency, and the contributions of individual components.

\subsection{Evaluation Protocol}
\label{sec:evaluation-protocol}

Our evaluation covers CALVIN ABC$\rightarrow$D~\cite{mees2022calvin},
ten RoboTwin~2.0 Hard tasks~\cite{chen2025robotwin2},
LIBERO~\cite{liu2023libero}, and six real-world manipulation tasks.
We report task success rates and, on CALVIN, the average number
of completed subtasks in a five-subtask sequence.

On LIBERO, we measure task success and command continuity
at handoff. The controlled continuation study uses
16-command chunks, an eight-command committed prefix,
and eight executed commands per handoff, with 50 paired
initial states for each of the 40 tasks.
We additionally compute tool-center-point (TCP) and joint
jerk RMS from simulated positions and joint angles over
the first five executed steps after handoff.

For execution efficiency, we report parameter counts,
action-expert evaluations per chunk, policy-response and
handoff latency, and the fraction of handoffs completed
within their deadlines at a fixed control rate.

\subsection{Main Benchmark Results}
\label{sec:original-benchmark-summary}

\begin{table}[!htbp]
\centering
\caption{Results on CALVIN ABC$\rightarrow$D.
Columns 1--5 report the percentage of chains completing at least
that many subtasks; Avg. len. is the mean completed sequence length.
$\dagger$ marks averages estimated from published completion rates.
B-VLA denotes BehaviorVLA.}
\label{tab:calvin_main}
\small
\setlength{\tabcolsep}{0pt}
\begin{tabular*}{\columnwidth}{@{\extracolsep{\fill}}lrrrrrr@{}}
\toprule
Method & 1 & 2 & 3 & 4 & 5 & Avg. $\uparrow$ \\
\midrule
RoboVLM & 98.0 & 93.6 & 85.4 & 77.8 & 70.4 & 4.25 \\
ReconVLA & 95.6 & 87.6 & 76.9 & 69.3 & 64.1 & $3.94^{\dagger}$\\
OpenVLA & 91.3 & 77.8 & 62.0 & 52.1 & 43.5 & 3.27 \\
UniVLA & 95.5 & 85.8 & 75.4 & 66.9 & 56.5 & 3.80 \\
UP-VLA & 92.8 & 86.5 & 81.5 & 76.9 & 69.9 & 4.08 \\
RoboDual & 94.4 & 82.7 & 72.1 & 62.4 & 54.4 & 3.66 \\
Seer & 96.3 & 91.6 & 86.1 & 80.3 & 74.0 & 4.28 \\
VPP & 95.7 & 91.2 & 86.3 & 81.0 & 75.0 & 4.29 \\
$\pi_0$~\cite{zhang2025dreamvla} & 93.8 & 85.0 & 76.7 & 68.1 & 59.9 & 3.84 \\
$\pi_{0.5}$~\cite{hu2026from} & 94.8 & 88.9 & 84.1 & 79.7 & 72.1 & $4.20^{\dagger}$\\
B-VLA & 96.0 & 92.0 & 87.3 & 82.9 & 77.3 & 4.36 \\
\textbf{Ours} & \textbf{98.9} & \textbf{95.4} & \textbf{91.0}
& \textbf{86.0} & \textbf{80.7} & \textbf{4.52} \\
\bottomrule
\end{tabular*}
\end{table}

\paragraph{Long-horizon composition on CALVIN.}
Table~\ref{tab:calvin_main} compares FluentAct with representative VLA
policies~\cite{li2024robovlm,song2025reconvla,kim2024openvla,
bu2025univla,zhang2025upvla,bu2024robodual,tian2024seer,
hu2024vpp,black2024pi0,black2025pi05}
and the behavior-aware baseline BehaviorVLA~\cite{hu2026from}.
FluentAct achieves the highest average completed sequence length
of 4.52, compared with 4.36 for BehaviorVLA.
It improves completion over BehaviorVLA by 2.9--3.7 percentage
points across all five chain depths, with five-subtask completion
increasing from 77.3\% to 80.7\%.

\begin{table}[!htbp]
\centering
\caption{Success rates (\%) on ten RoboTwin~2.0 Hard tasks.
Avg. is the mean over the first five tasks; Overall is the mean
over all ten. B-VLA denotes BehaviorVLA.}
\label{tab:robotwin_main}
\small
\setlength{\tabcolsep}{0pt}
\begin{tabular*}{\columnwidth}{@{\extracolsep{\fill}}lrrrrrr@{}}
\toprule
Method & AB & CA & CB & DB & GR & Avg.\ I \\
\midrule
RDT & 75.0 & 12.0 & 9.0 & 32.0 & 43.0 & 34.2 \\
ACT & 23.0 & 4.0 & 3.0 & 1.0 & 25.0 & 11.2 \\
DP & 0.0 & 5.0 & 0.0 & 0.0 & 0.0 & 1.0 \\
DP3 & 3.0 & 14.0 & 0.0 & 53.0 & 2.0 & 14.4 \\
$\pi_0$ & 56.0 & 11.0 & 3.0 & 24.0 & 80.0 & 34.8 \\
$\pi_{0.5}$ & 75.0 & 44.0 & 64.0 & 69.0 & 82.0 & 66.8 \\
B-VLA & 83.0 & 52.0 & 77.0 & 77.0 & 90.0 & 75.8 \\
\textbf{FluentAct} & \textbf{88.0} & \textbf{56.0} & \textbf{82.0} & \textbf{83.0} & \textbf{94.0} & \textbf{80.6} \\
\midrule
Method & MP & P2R & PB & BF & CP & Overall \\
\midrule
RDT & 11.0 & 1.0 & 2.0 & 27.0 & 17.0 & 22.9 \\
ACT & 0.0 & 0.0 & 0.0 & 0.0 & 1.0 & 5.7 \\
DP & 0.0 & 0.0 & 0.0 & 0.0 & 0.0 & 0.5 \\
DP3 & 3.0 & 0.0 & 1.0 & 18.0 & 1.0 & 9.5 \\
$\pi_0$ & 22.0 & 6.0 & 4.0 & 4.0 & 45.0 & 25.5 \\
$\pi_{0.5}$ & 32.0 & 19.0 & 28.0 & 46.0 & 55.0 & 51.4 \\
B-VLA & 41.0 & 25.0 & 36.0 & 61.0 & 62.0 & 60.4 \\
\textbf{FluentAct} & \textbf{49.0} & \textbf{30.0} & \textbf{40.0} & \textbf{67.0} & \textbf{69.0} & \textbf{65.8} \\
\bottomrule
\end{tabular*}
\end{table}

\paragraph{Bimanual manipulation on RoboTwin.}
Table~\ref{tab:robotwin_main} compares FluentAct with
RDT, ACT, Diffusion Policy, DP3, $\pi_0$, $\pi_{0.5}$, and
BehaviorVLA~\cite{liu2024rdt,zhao2023act,chi2023diffusionpolicy,
ze2024dp3,black2024pi0,black2025pi05,hu2026from}.
FluentAct achieves 65.8\% mean success, compared with 60.4\%
for BehaviorVLA, a gain of 5.4 percentage points.
Success improves on all ten tasks by 4--8 percentage points.
The largest gain is on Move Playingcard Away,
where success increases from 41.0\% to 49.0\%.

\begin{figure*}[!t]
\centering
\includegraphics[width=\textwidth]{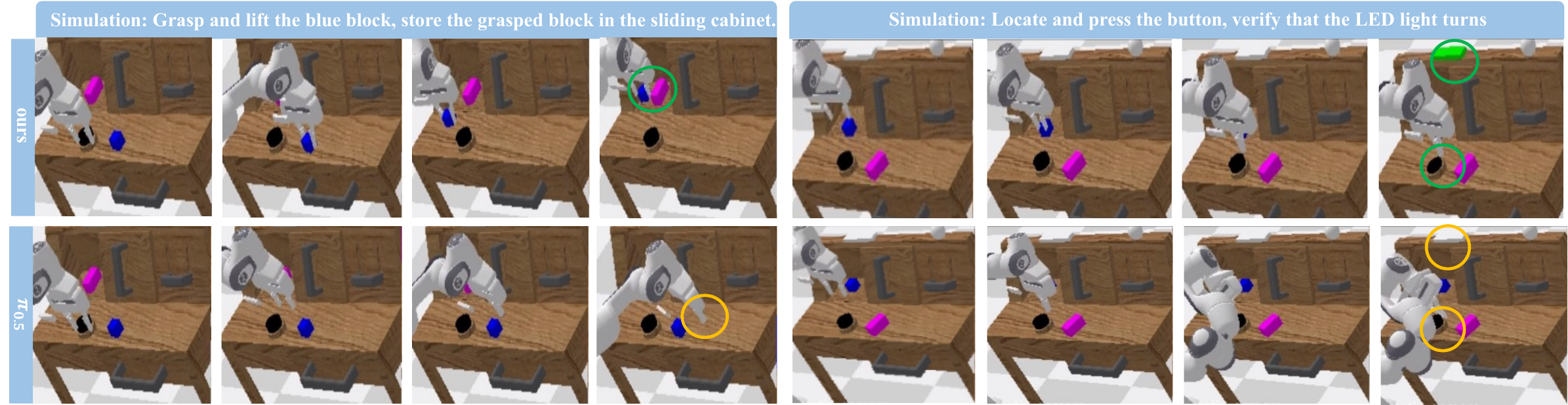}
\caption{Qualitative comparison in simulation.
FluentAct (top) and $\pi_{0.5}$ (bottom) perform block storage
and button activation. Circles highlight the final outcomes.}
\label{fig:simulation}
\end{figure*}

\paragraph{Qualitative simulation comparison.}
Figure~\ref{fig:simulation} shows block storage and button activation.
FluentAct completes both tasks in the illustrated rollouts,
whereas $\pi_{0.5}$ leaves the block outside the cabinet
and fails to activate the indicator light.

\begin{figure*}[!t]
\centering
\includegraphics[width=\textwidth]{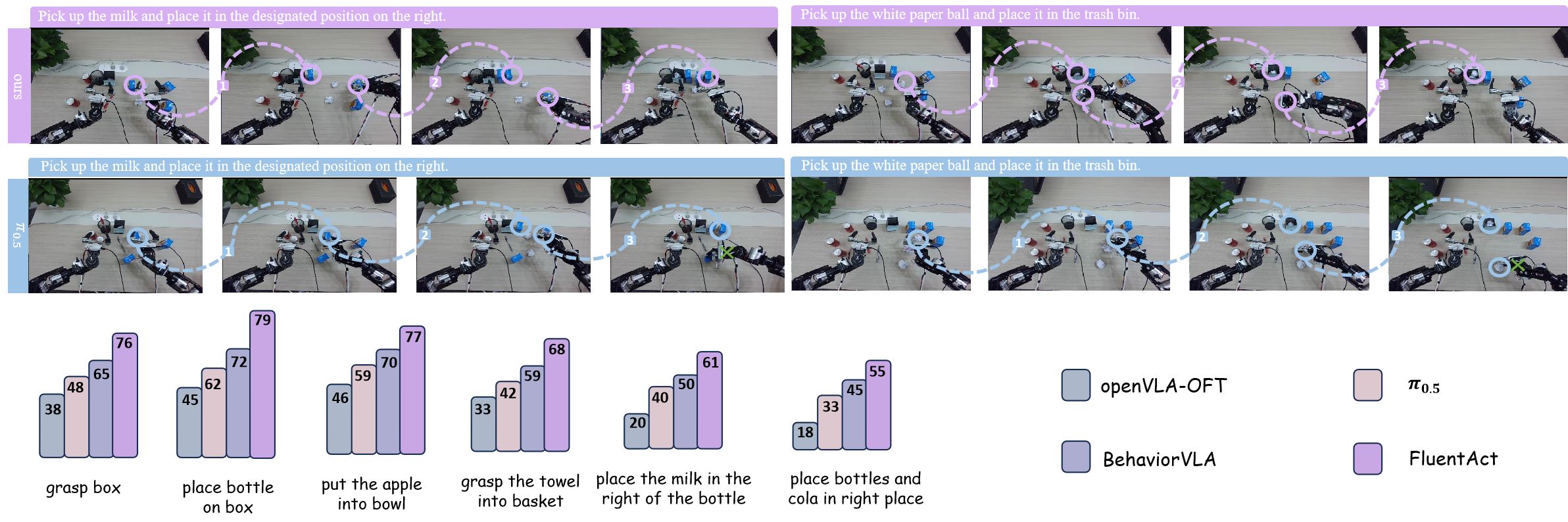}
\caption{Real-world evaluation.
Top: example rollouts of FluentAct and $\pi_{0.5}$ on milk
placement and paper-ball disposal.
Bottom: success rates (\%) on six manipulation tasks.}
\label{fig:realworld}
\end{figure*}

\subsection{Real-World Evaluation}
\label{sec:real-world}

We evaluate FluentAct on six real-world manipulation tasks
using a dual-arm robot. Figure~\ref{fig:realworld} reports
per-task success alongside example rollouts.
FluentAct achieves 69.3\% mean success, compared with 60.2\%
for BehaviorVLA, 47.3\% for $\pi_{0.5}$, and 33.3\%
for OpenVLA-OFT. It outperforms BehaviorVLA on all six
tasks by 7--11 percentage points.

\subsection{Continuation Preference Learning}
\label{sec:continuation-evaluation}

\begin{table}[!htbp]
\centering
\caption{Effect of continuation preference learning on LIBERO.
SR denotes task success rate. Jump and Trend are averaged over
handoffs in de-normalized command space.
Chosen-only SFT regresses preferred continuations;
Continuity DPO uses the reference-relative preference objective.}
\label{tab:libero_bicpo}
\small
\setlength{\tabcolsep}{0pt}
\begin{tabular*}{\columnwidth}{@{\extracolsep{\fill}}lrrr@{}}
\toprule
Method & SR (\%)$\uparrow$ & Jump$\downarrow$ & Trend$\downarrow$ \\
\midrule
\textbf{FluentAct} & \textbf{99.1} & \textbf{2.37} & \textbf{6.40} \\
\quad w/o DPO & 98.8 & 3.01 & 8.07 \\
\quad Chosen-only SFT & 97.9 & 2.85 & 6.80 \\
\midrule
$\pi_{0.5}$-FM & 96.9 & 4.22 & 8.50 \\
\quad Chosen-only SFT & 49.1 & 3.13 & 6.90 \\
\quad Continuity DPO & \textbf{97.1} & \textbf{2.52} & \textbf{6.70} \\
\midrule
Legato & 97.5 & 2.98 & 7.60 \\
\quad Chosen-only SFT & 96.6 & 2.63 & 7.10 \\
\quad Continuity DPO & \textbf{97.6} & \textbf{2.51} & \textbf{6.60} \\
\midrule
RTC & 97.3 & 3.07 & 7.80 \\
\quad Chosen-only SFT & 95.4 & 2.65 & 7.30 \\
\quad Continuity DPO & \textbf{97.4} & \textbf{2.58} & \textbf{6.90} \\
\bottomrule
\end{tabular*}
\end{table}

\paragraph{LIBERO handoff quality and objective transfer.}
Table~\ref{tab:libero_bicpo} reports LIBERO success and handoff
continuity. For each handoff, we report
$\mathrm{Jump}=\|a_0-u_{-1}\|_2$
and
$\mathrm{Trend}=\|(a_1-a_0)-(u_{-1}-u_{-2})\|_2$.
Both are unweighted $\ell_2$ distances computed in
de-normalized command space and averaged over handoffs.
The training trend cost in Eq.~\eqref{eq:continuation-ranking}
uses the boundary increment $a_0-u_{-1}$ in place of
the first incoming increment $a_1-a_0$.
Compared with the variant without preference training, FluentAct
reduces jump by 21.3\% and trend mismatch by 20.7\%, with 99.1\%
task success.

We apply the reference-relative pairwise objective to
$\pi_{0.5}$-FM, Legato~\cite{liu2026learning}, and
RTC~\cite{black2025real} using each policy's native conditional
flow-matching energy and retaining its generator and conditioning.
Jump decreases by 15.8--40.3\% and trend mismatch by 11.5--21.2\%
across these policies. For $\pi_{0.5}$-FM, the chosen-only control retains
preferred-candidate regression and demonstration replay,
with the same initialization, training budget, and
optimization settings. Only the pairwise preference
term is removed. Preference optimization also achieves
lower handoff metrics and higher success than chosen-only SFT
for all four policies.

\paragraph{Controlled continuation study.}
We compare FluentAct, Haar imitation, and extra imitation
across three training seeds with paired initial states.
FluentAct and extra imitation start from the same Haar
imitation checkpoint and receive 30,000 additional updates
with the same batch size.
Averaged over the three seeds, FluentAct achieves higher
mean task success across the four LIBERO suites than both
imitation controls. On LIBERO-Long, it also achieves lower
mean TCP and joint jerk RMS over the first five executed
steps after handoff than Haar imitation.
The supplementary material reports per-seed results,
cross-seed means and standard deviations, and the detailed
evaluation protocol.

\subsection{Inference Delay and Execution Efficiency}
\label{sec:delay-robustness}

\begin{table}[!htbp]
\centering
\caption{Effect of inference delay on LIBERO.
Delays are measured in control steps.
The random-delay setting samples $k$ from $\{3,4,5\}$.}
\label{tab:delay_results}
\small
\setlength{\tabcolsep}{0pt}
\begin{tabular*}{\columnwidth}{@{\extracolsep{\fill}}lrrrr@{}}
\toprule
Metric & $k=3$ & $k=4$ & $k=5$ & Random \\
\midrule
SR (\%) $\uparrow$ & \textbf{99.1} & 98.9 & 98.8 & 98.8 \\
Jump$\downarrow$ & \textbf{2.37} & 2.47 & 2.50 & 2.46 \\
Trend$\downarrow$ & 6.40 & 6.21 & 6.23 & \textbf{6.17} \\
\bottomrule
\end{tabular*}
\end{table}

Table~\ref{tab:delay_results} evaluates fixed delays of three,
four, and five control steps and random delays sampled from
the same set. Success ranges from 98.8\% to 99.1\%.
With fixed delays, command jump increases from 2.37 at $k=3$
to 2.50 at $k=5$. Trend mismatch ranges from 6.17 to 6.40
across all four settings.

\begin{table}[!htbp]
\centering
\caption{Model size and inference efficiency in unpaced LIBERO
simulation on an RTX~4090.
Fwd./VJP counts action-expert forward evaluations and action-input
vector--Jacobian products per chunk.
We report p50 and p95 latencies (ms) over 30 matched requests
after warm-up.}
\label{tab:handoff-timing}
\small
\setlength{\tabcolsep}{0pt}
\begin{tabular*}{\columnwidth}
{@{\extracolsep{\fill}}lrrrrrr@{}}
\toprule
& & & \multicolumn{2}{c}{Response}
    & \multicolumn{2}{c}{Handoff} \\
\cmidrule(lr){4-5}\cmidrule(l){6-7}
Method & Params. (B) & Fwd./VJP
       & p50 & p95 & p50 & p95 \\
\midrule
Native $\pi_{0.5}$ & 3.617 & 10/0
    & 223 & 255 & 227 & 258 \\
RTC & 3.617 & 10/10
    & 544 & 647 & 553 & 656 \\
Haar imitation & 3.670 & 2/0
    & \textbf{130} & \textbf{159}
    & \textbf{143} & \textbf{171} \\
FluentAct & 3.670 & 2/0
    & 135 & 172 & 153 & 189 \\
\bottomrule
\end{tabular*}
\end{table}

\paragraph{Request-to-handoff latency.}
We measure thirty warmed requests per policy across three
LIBERO contexts on one RTX~4090 at batch size one.
All methods use sixteen-command chunks, eight-command
committed prefixes, and eight-command execution intervals.
Requests are matched by task, initial state, and request index
in the unpaced simulator.
Table~\ref{tab:handoff-timing} reports elapsed time from request
submission to receipt of the policy response and to execution
of the first incoming command.
FluentAct achieves a median handoff latency of 153\,ms,
32.5\% below native $\pi_{0.5}$ and 72.2\% below RTC.
FluentAct and Haar imitation use two action-expert forward
evaluations per chunk, compared with ten for native $\pi_{0.5}$.
RTC requires ten forward evaluations and ten action-input VJPs.
FluentAct has 1.47\% more parameters than the native policy.

\paragraph{Handoffs at a fixed control rate.}
In three 64-step windows at 20 Hz on an RTX 4090,
FluentAct and native $\pi_{0.5}$ meet all 21 handoff
deadlines under a 400 ms budget. RTC misses the first
deadline in each window, ending each run.

\subsection{Ablation Studies}
\label{sec:ablations}

\begin{table}[!htbp]
\centering
\caption{Component ablation on CALVIN ABC$\rightarrow$D.
The variant without Haar predicts in the original action coordinates.
Columns 1--5 report chain-completion rates (\%);
Avg. len. is the mean completed sequence length.}
\label{tab:component_ablation}
\small
\setlength{\tabcolsep}{0pt}
\begin{tabular*}{\columnwidth}{@{\extracolsep{\fill}}lrrrrrr@{}}
\toprule
Variant & 1 & 2 & 3 & 4 & 5 & Avg. $\uparrow$ \\
\midrule
\textbf{Full} & \textbf{98.9} & \textbf{95.4} & \textbf{91.0}
& \textbf{86.0} & \textbf{80.7} & \textbf{4.520} \\
w/o Haar & 98.2 & 94.8 & 90.7 & 85.2 & 79.6 & 4.485 \\
w/o preference & 97.8 & 95.0 & 90.5 & 85.4 & 78.9 & 4.476 \\
w/o instruction & 97.0 & 92.5 & 87.9 & 84.0 & 78.2 & 4.396 \\
\bottomrule
\end{tabular*}
\end{table}

\paragraph{Component ablations.}
Table~\ref{tab:component_ablation} reports component ablations on CALVIN.
Removing instruction conditioning from behavior encoding
reduces the average completed sequence length from 4.520
to 4.396 and five-subtask completion from 80.7\% to 78.2\%.
Five-subtask completion is 78.9\% without continuation
preference learning and 79.6\% when the generator predicts
in the original action coordinates.

\paragraph{Sampling steps.}
With fixed seed-44 weights on 500 paired LIBERO-Long
initial states, one Euler step per stage achieves
82.0\% success versus 83.6\% with two.
In a separate A800 timing test, one step reduces median
policy inference latency by 22.5\% relative to two.
The full comparison of one, two, and four steps per
stage is reported in the supplementary material.

\paragraph{Location of instruction conditioning.}
Table~\ref{tab:instruction_ablation} compares three locations
for instruction conditioning with matched capacity.
Conditioning visual queries before history aggregation
achieves an average completed sequence length of 4.520,
compared with 4.476 for query fusion and 4.417 for memory fusion.
Five-subtask completion is 80.7\%, 80.3\%, and 78.6\%,
respectively.

\begin{table}[!htbp]
\centering
\caption{Effect of instruction conditioning on CALVIN ABC$\rightarrow$D.
The variants use different conditioning locations with matched capacity.
Columns 1--5 report chain-completion rates (\%);
Avg. len. is the mean completed sequence length.}
\label{tab:instruction_ablation}
\small
\setlength{\tabcolsep}{0pt}
\begin{tabular*}{\columnwidth}{@{\extracolsep{\fill}}lrrrrrr@{}}
\toprule
Variant & 1 & 2 & 3 & 4 & 5 & Avg. $\uparrow$ \\
\midrule
Visual query & \textbf{98.9} & \textbf{95.4} & \textbf{91.0} & \textbf{86.0} & \textbf{80.7} & \textbf{4.520} \\
Query fusion & 98.6 & 94.4 & 89.2 & 85.1 & 80.3 & 4.476 \\
Memory fusion & 97.5 & 93.1 & 88.3 & 84.2 & 78.6 & 4.417 \\
\bottomrule
\end{tabular*}
\end{table}

%% file: sec/5_conclusion.tex
\section{Conclusion}
\label{sec:conclusion}

We presented FluentAct, a framework for asynchronous
vision-language-action control that integrates instruction-aware
behavior context, handoff-conditioned ordered Haar generation,
and offline continuation preference learning.
FluentAct improves task completion over BehaviorVLA on CALVIN,
RoboTwin~2.0 Hard, and six real-world manipulation tasks.
On LIBERO, the reference-relative preference objective reduces
command jump and trend mismatch at handoff for FluentAct,
$\pi_{0.5}$-FM, Legato, and RTC.
At deployment, FluentAct generates each continuation with
two forward evaluations of a shared action expert and no
online candidate ranking. Sampling ablations show a favorable success--latency tradeoff.
In unpaced LIBERO simulation on an RTX~4090, its median
request-to-handoff latency is 153\,ms, 32.5\% lower than
native $\pi_{0.5}$.